\documentclass[a4paper,fleqn]{cas-dc}  %sc, dc

\usepackage{float}  % Force figure to be set at a page
\usepackage{svg}  % for figure format svg
\usepackage[square,numbers,compress]{natbib} % citation package
\usepackage{float}  % 'H' option for figure
\begin{document}
\let\WriteBookmarks\relax
\def\floatpagepagefraction{1}
\def\textpagefraction{.001}

\renewcommand{\printorcid}{} % This is to remove orcid 

%\linenumbers % Enable line numbering
% Main title of the paper

%\fontsize{8.5}{11}\selectfont
% Short title
%\shorttitle{}
\shorttitle{Data-Driven Assessment of Concrete Mixture Compositions on Chloride Transport via Standalone Machine Learning Algorithms}    

% Short author
\shortauthors{Aliasghar-Mamaghani and Khalafi}

\title [mode = title]{Data-Driven Assessment of Concrete Mixture Compositions on Chloride Transport via Standalone Machine Learning Algorithms}  

\author[1]{Mojtaba Aliasghar-Mamaghani} %

% Corresponding author indication
\cormark[1]
% Footnote of the first author
%\fnmark[]  % this would enable note 1 in title
% Email id of the first author
\ead{mojtaba@austin.utexas.edu}
%Credit authorship
\credit{Writing – review \& editing, Writing – original draft, Software, Methodology, Investigation, Formal analysis, Conceptualization}
%\credit{Experiment, Analysis}
% Address/affiliation
\affiliation[1]{position={Postdoctoral Fellow,},
            department={Department of Civil Architectural and Environmental Engineering,},
            organization={The University of Texas at Austin},
            city={Austin},
            state={TX},
            country={United States}}

\author[2]{Mohammadreza Khalafi}
\affiliation[2]{position={Student,},
            department={Department of Civil Engineering,},
            organization={Sharif University of Technology},
            city={Tehran},
            country={Iran}}

\ead{mohammad.khalafi85@sharif.edu}
% Credit authorship
\credit{Writing – review \& editing, Writing – original draft, Software, Methodology, Investigation, Formal analysis, Conceptualization}
% Corresponding author text
\cortext[1]{Corresponding author} %% \vspace{-10pt}

% Footnote text
%\fntext[1]{}

% For a title note without a number/mark
%\nonumnote{}

% Here goes the abstract
\begin{abstract}
This paper employs a data-driven approach to determine the impact of concrete mixture compositions on the temporal evolution of chloride in concrete structures. This is critical for assessing the service life of civil infrastructure subjected to aggressive environments. The adopted methodology relies on several simple and complex standalone machine learning (ML) algorithms, with the primary objective of establishing confidence in the unbiased prediction of the underlying hidden correlations. The simple algorithms include linear regression (LR), k-nearest neighbors (KNN) regression, and kernel ridge regression (KRR). The complex algorithms entail support vector regression (SVR), Gaussian process regression (GPR), and two families of artificial neural networks, including a feedforward network (multilayer perceptron, MLP) and a gated recurrent unit (GRU). The MLP architecture cannot explicitly handle sequential data, a limitation addressed by the GRU. A comprehensive dataset is considered. The performance of ML algorithms is evaluated, with KRR, GPR, and MLP exhibiting high accuracy. Given the diversity of the adopted concrete mixture proportions, the GRU was unable to accurately reproduce the response in the test set. Further analyses elucidate the contributions of mixture compositions to the temporal evolution of chloride. The results obtained from the GPR model unravel latent correlations through clear and explainable trends. The MLP, SVR, and KRR also provide acceptable estimates of the overall trends. The majority of mixture components exhibit an inverse relation with chloride content, while a few components demonstrate a direct correlation. These findings highlight the potential of surrogate approaches for describing the physical processes involved in chloride ingress and the associated correlations, toward the ultimate goal of enhancing the service life of civil infrastructure.

%\nocite{*}%% Remove this line from your manuscript.
\end{abstract}

% Use if graphical abstract is present
%\begin{graphicalabstract}
%\includegraphics{}
%\end{graphicalabstract}

% Research highlights
%\begin{highlights}
%\item 
%Large-scale experimental test conducted on a %non-ductile reinforced concrete wall.
%\end{highlights}

%\nocite{*}

% Keywords
% Each keyword is seperated by \sep

\maketitle

% Main text
\section{Introduction}\label{Introduction}
Corrosion is one of the predominant causes of long-term deterioration of civil infrastructure components. Concrete bridges are commonly employed construction systems within the daily commute networks. During the early stages of concrete setting, a protective film forms around the reinforcing materials due to the high alkalinity of the concrete medium. The formation of this protective layer places the reinforcing materials in a passive state, effectively reducing the risk of corrosion. Electrochemical oxidation-reduction reactions characterize the propensity for corrosion. The breakdown of the protective film leads to a substantial increase in the rate of corrosion reactions, a critical point after which significant serviceability issues occur~\cite{bavzant1979physical,hansen1999numerical,weyers1994concrete,angst2019predicting,aliasghar2023computational}.

Corrosion imposes several deleterious effects on concrete structures. In particular, it adversely affects the mechanical characteristics and deformability of the reinforcing materials, as well as the bond properties between steel and concrete~\cite{rodriguez1997load,cairns2005mechanical,pape2011effects,solgaard2013concrete, vecchi2021corrosion, habibi2022modeling, aliasghar2022analytical}. Furthermore, the formation of corrosive products—--which often exhibit a volumetric expansion 2 to 6.4 times greater than the original reactants~\cite{youping1996modeling}—--induces substantial passive pressure in the surrounding concrete, leading to the development of tensile strains and, subsequently, deterioration or spalling in affected regions~\cite{hansen1999numericalb,murray1992chloride,aliasghar2025multiphysics}.

The serviceability of concrete structures is profoundly affected by the content of chloride at the level of the reinforcing materials. Chloride ions destroy the protective layer initially formed around the steel reinforcement, thereby depassivating the steel, and inducing a substantial increase in the corrosion rate~\cite{bavzant1979physical,maekawa2003multi,maekawa2008multi,hussain2011enhanced,aliasghar2023computational}. 
Several prior studies have proposed replacing conventional steel reinforcement with alternative materials to circumvent the serviceability issues associated with steel corrosion~\cite{bazli2016effect,aliasghar2019seismic,aliasghar2021effective,hassanpour2022effect}. Despite the significant efforts made, reinforced and prestressed concrete structures remain predominant worldwide~\cite{aliasghar2022analytical, tran2015cyclic, aliasghar2024experimental,aliasghar2024experimentalb}.

Several previous studies have focused on the description of chloride ingress in the concrete pore network. The intrusion of chloride in concrete can be simply described by a closed-form equation on the basis of Fick's law of diffusion~\cite{hansen1999numerical,collepardi1972penetration}. Advanced finite element approaches have been proposed to capture several interacting phenomena involved in chloride ingress~\cite{saetta1993analysis,maekawa2003multi, maekawa2008multi,aliasghar2023finite, aliasghar2023computational}. The potential caveat of such numerical methodologies pertains to the several interacting physical mechanisms inherent to the problem setting, a coupled procedure that often entails substantial computational expense~\cite{aliasghar2023finite, aliasghar2023computational}.

In light of the aforementioned limitations, several studies have endeavored to develop surrogate frameworks on the basis of Machine Learning (ML) algorithms to describe the evolution of chloride in concrete. The primary motivation for such efforts lies in the considerable computational effort associated with physics-based modeling schemes, with the ultimate goal of delivering a cost-effective approach~\cite{cai2020prediction,taffese2022machine, hosseinzadeh2023efficient,zheng2024optimum,tran2022machine,shaban2023physics, aliasghar2025quantitative}.

Various approaches have been adopted in the past. Simplified approaches have primarily focused on quantifying the chloride diffusion coefficient involved in the closed-form solution~\cite{hosseinzadeh2023efficient,cai2020prediction,tran2022machine,taffese2022machine,zheng2024optimum}. The capability of the physics-informed approaches to reproduce chloride content has also been investigated~\cite{shaban2023physics}. Recent advances have further shown the ability of several ML algorithms to accurately reproduce the chloride content in concrete while concurrently accounting for various interacting phenomena~\cite{aliasghar2025quantitative}.

The vast majority of previous efforts have focused on an isolated facet of the problem, with particular emphasis on determining the diffusivity coefficient entailed in the closed-form solution. Considering the complexity inherent in the problem setting, the temporal evolution of chloride has been rarely investigated through ML techniques. The correlation between concrete mixture compositions and the temporal evolution of chloride is another important aspect that has received limited attention. This is of significant importance, as it directly characterizes the service life of civil infrastructure in extreme environments. The need for such capabilities has become increasingly urgent considering the growing concerns pertaining to the impacts of climate change and structural aging on the service life of structures in harsh environments. The availability of such knowledge substantially reinforces the understanding required for optimizing concrete mixture properties and, subsequently, for the design of resilient and long-lasting civil infrastructure.

In order to address the foregoing research need, this study employs various standalone ML algorithms to describe the temporal evolution of chloride in concrete. The methodology incorporates three simple ML algorithms, including linear regression, k-nearest neighbors regression, and kernel ridge regression, as well as four more complex algorithms, including support vector regression, Gaussian process regressor, and two families of artificial neural networks encompassing a feedforward network (i.e., multilayer perceptron, MLP), and a gated recurrent unit. The latter was particularly employed to investigate the sequential response. A wide range of concrete mixture properties, along with various environmental factors, is included in the dataset. The accuracy of the algorithms is evaluated. Further analyses elucidate the contributions of mixture compositions to the temporal evolution of chloride and, subsequently, the role of these features in enhancing the service life of civil infrastructure in harsh environments.

\section{Description of Algorithms}\label{Cmp-Fmw}
%\vspace{0.5em}  
The development of the models is driven by a variety of standalone ML algorithms, including three relatively simple approaches and four more complex methods, with the objective of identifying the hidden correlations between the temporal evolution of chloride and concrete mixture properties. The primary motivation for employing multiple ML algorithms is to evaluate the capabilities of each model and to establish confidence in the unbiased prediction of the underlying hidden correlations.

The first class of algorithms is the linear regression (LR) method. The algorithm provides a simple estimate of the target variable by establishing a linear relation between the input features and the output. The model entails parameters that are estimated using the least squares method by minimizing the sum of squared residuals~\cite{friedman2009elements}.

The second simple type of algorithm considered rely on k-nearest neighbors (KNN) regression. The specific method is a non-parametric, instance-based learning regressor that makes use of \textit{k} points (closest neighbors) in the feature space to predict the output. In this study, the prediction is obtained using a weighted average of these neighbors~\cite{fix1985discriminatory,friedman2009elements}.

Kernel ridge regression (KRR) is another ML algorithm adopted in this study. The method builds on ridge regression and makes use of kernel trick to implicitly map input features into a higher-dimensional space. The ridge component introduces a penalty parameter to prevent overfitting and improve generalization by shrinking the coefficients~\cite{tibshirani1996regression,friedman2009elements}.

Support vector regression (SVR) is a commonly employed ML algorithm across a variety of fields in artificial intelligence. Kernel ridge regression (KRR) and SVR are both grounded in kernel-based learning frameworks, wherein a kernel function is adopted to establish correlations in high-dimensional feature space; however, their respective loss functions and optimization objectives are different. The SVR formulation adopted in this study is the~$\epsilon$-SVR model, which permits errors to reside within the $\epsilon$-insensitive tube while penalizing deviations outside this region~\cite{pedregosa2011scikit, haykin2009neural, smola2004tutorial}.

Gaussian process regression (GPR) is another class of algorithms adopted in this study, a nonparametric Bayesian model, which also employs the kernel trick to characterize underlying patterns. The development of the model is driven by the assumption that the underlying function values, evaluated across the feature space, follow a joint Gaussian distribution. GPR further incorporates an independent noise term in the predictions, which is modeled as a Gaussian random variable, typically with zero mean. The kernel function employed in this study is the radial basis function~\cite{rasmussen2003gaussian,schulz2018tutorial}.

The final class of algorithms relies on artificial neural networks (ANNs). ANNs were initially inspired by biological neural networks, with the objective of reproducing human-like performance~\cite{jain1996artificial, lippmann1988introduction}. The first type of ANN considered in this study is the multilayer perceptron (MLP), which aims to capture the rather complex relation between the input and output features. The MLP is a subclass of feed-forward neural networks, with an architecture that enables the information to evolve in a forward direction. The architecture makes use of multiple layers, each with numerous neurons interconnected through synaptic, weighted links. Each neuron contains a computational unit that receives inputs from the preceding layer and subsequently delivers an output, typically following an operation through an activation function. The training of the algorithm is commonly conducted through a back-propagation procedure that employs a gradient-descent scheme for optimization. The latter is performed on the weight factors associated with the synaptic links so as to enhance the efficiency of the network.

The architecture of the commonly-employed MLP algorithms is inherently devoid of the capability to explicitly handle sequential modeling. Therefore, the final advanced ML algorithm adopted in this study relies on the Gated Recurrent Unit (GRU)~\cite{cho2014properties}. The GRU is a subclass of Recurrent Neural Networks (RNNs) with gating mechanisms that adaptively capture temporal dependencies, while mitigating some of the caveats of the traditional RNNs, such as the vanishing gradient problem. The motivation for employing GRUs in this study stems from the inherent temporal dependencies present in the dataset.

\section{Dataset and Model Development}\label{dataset-framework}
%\vspace{0.5em}  
The dataset adopted in this study relies on the experimental tests conducted on concrete specimens in the literature~\cite{sergi1992diffusion, maruya1998modeling, luping2003chloride,maekawa2003multi,samson2007modeling, baroghel2014performance, kim2016chloride, pradelle2016comparison, chen2021effect} and is available through an online repository~\cite{aliasghar2025quantitativeData}. The present study primarily focuses on diffusion and neglects advection. In this respect, the dataset is devoid of any experimental instance with a driving force governed by moisture flow (advection). The implication of this assumption on the existing dataset~\cite{aliasghar2025quantitativeData, aliasghar2025quantitative} was to exclude moisture-driven data (advection) and solely focus on concentration-gradient-driven data (diffusion). This assumption was pursued to enhance the reliability of results and forgo any potential error that might arise from advection.

The analyzed dataset contains a range of environmental factors and mixture properties that constitute the inventory of input and output variables. Within the context of this study, the quantity of interest is the chloride content in the porous network of concrete. The inventory of input variables includes environmental factors such as the content of surface chloride, time of exposure, temperature, as well as depth from the exposure surface, concomitant with concrete mixture compositions, including the content of water, sulfate-resisting Portland cement (SPRC), ordinary Portland cement (OPC), water-to-binder ratio, fly ash (FA), silica fume (SF), ground granulated blast-furnace slag (GGBS), superplasticizer, and fine and coarse aggregates. 

The development of the model and the implementation of the framework relied on established libraries in the literature~\cite{pedregosa2011scikit, paszke2019pytorch}. The process involved in training included the separation of the dataset into training and test splits, with the respective ratios of 0.75 and 0.25. The dataset was then normalized to have a mean value of 0 and a standard deviation of 1.0. Subsequently, the fine-tuning process employed 10-fold cross-validation to mitigate the potential for overfitting. 

The performance of the ML algorithms was appraised employing a variety of metrics, including the coefficient of determination ($R^2$), mean absolute error (MAE), mean square error (MSE) and its root (RMSE), and mean absolute percentage error (MAPE). The best performance is obtained when the coefficient of determination approaches unity, while values approaching zero indicate the best results for the other metrics.

\section{Results}\label{Results}
\begin{figure*}[]
    \centering    
    \includegraphics[width=1.0\textwidth]{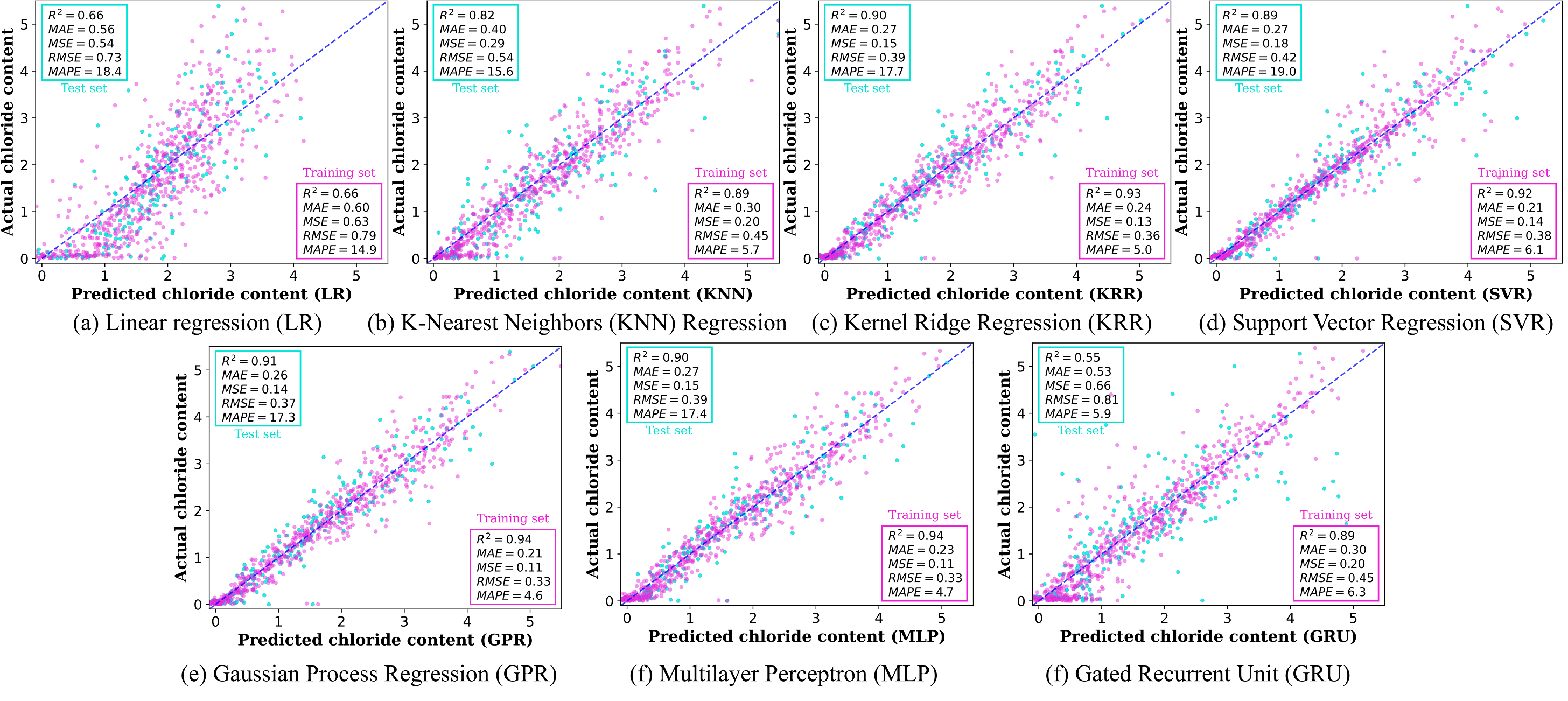}
    \vspace{-5mm}
    \caption{Evaluation metrics for the adopted ML algorithms}
    \label{Fig1-Results}
    \vspace{-2mm}
\end{figure*}

%\vspace{0.5em} 
The prediction capability of each ML algorithm is summarized in Figure~\ref{Fig1-Results}. Considering the characteristics of each learner, as anticipated, linear regression (LR) provides a poor estimate, with a performance metric of $R^2=0.62$. The k-nearest neighbors (KNN) regression demonstrates a relatively satisfactory estimate, achieving the respective $R^2$ scores of $0.89$ and $0.82$ for the training and test splits. 

The kernel ridge regression (KRR) and support vector regression (SVR) exhibit good predictive performance with the respective test-set $R^2$ scores of 0.90 and 0.89. The Gaussian process regression (GPR) provides an accurate estimate, exhibiting a test-set score of $R^2 = 0.91$. 

Subsequently, considering the adopted artificial neural networks, the multilayer perceptron (MLP) provides good performance, with a test-set score of $R^2=0.90$. On the other hand, the Gated recurrent unit (GRU) shows poor performance, achieving a score of $R^2=0.55$.

Pursuant to the establishment of models and the evaluation of their accuracy in predicting the content of chloride, further investigations are conducted on the sensitivity of the concrete mixture compositions to the temporal evolution of chloride. In light of the foregoing evaluation metrics and the accuracy achieved, the GRU was delisted from the inventory of ML learners for subsequent analyses.

The impact of concrete mixture properties on the temporal evolution of chloride in concrete is elucidated by examining the correlations between concrete mixture constituents and the space-time domain. The investigation is carried out using a reference concrete mixture as a baseline~\cite{kim2016chloride}, from which the influence of individual mixture properties is appraised by varying each parameter over its entire range observed in the dataset~\cite{aliasghar2025quantitativeData}. 
The reference concrete mixture consisted of 184~kg/m$^3$ of water, 460~kg/m$^3$ of Ordinary Portland cement, 100~kg/m$^3$ of fly ash, and fine and coarse aggregate contents of 700 and 1050~kg/m$^3$, respectively, together with superplasticizer content of 1.8~kg/m$^3$. The specimen was exposed to a surface chloride of 19.6~gr/l for a duration of 1.3~years under an average ambient temperature of 9~$^\circ$C~\cite{kim2016chloride}. The reference point for the temporal plot was fixed at a location 10~mm away from the exposure surface---unless otherwise specified.

\begin{figure*}[]
    \centering        
    \includegraphics[width=0.95\textwidth]{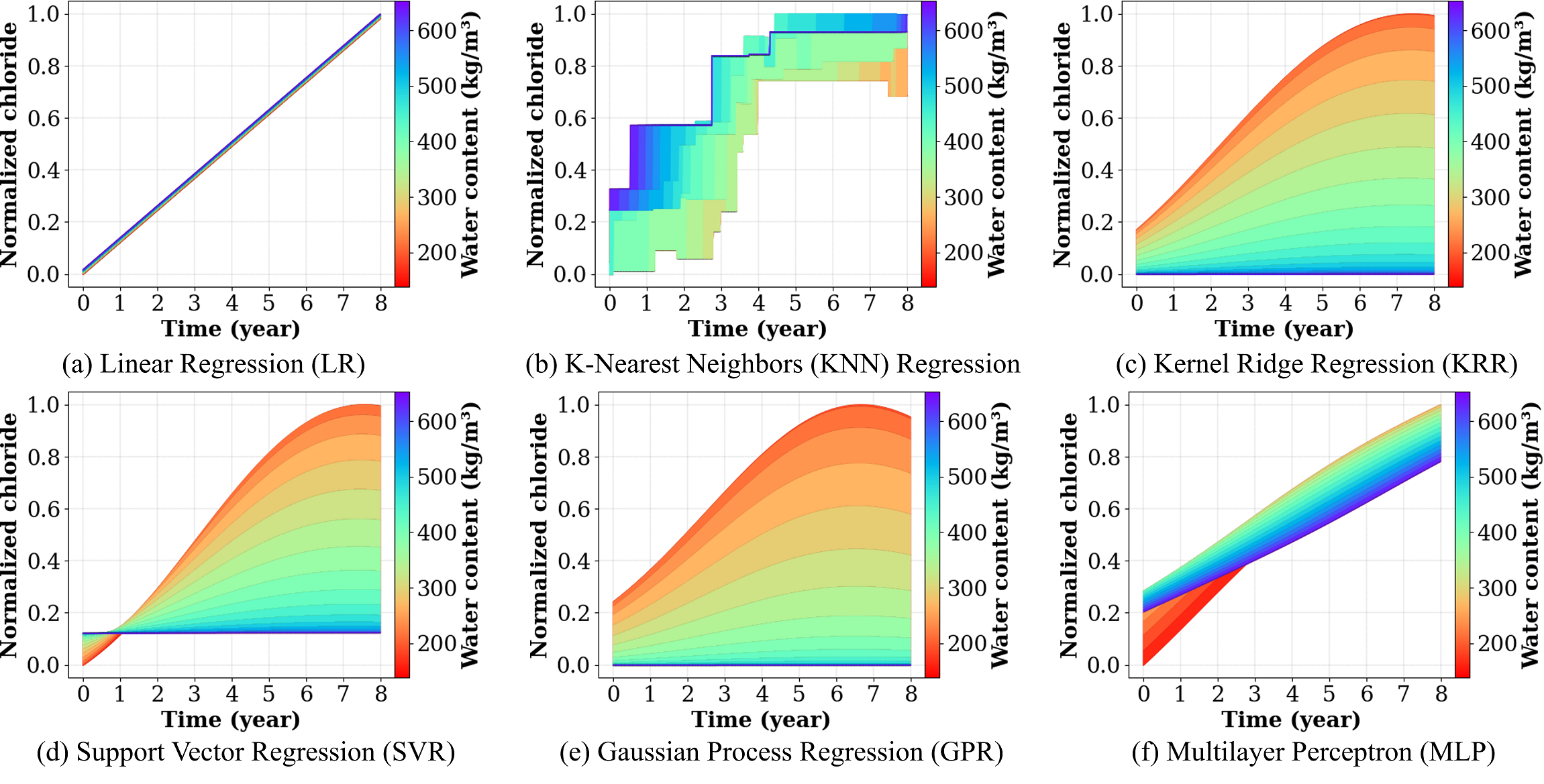}
    \vspace{-5mm}
    \caption{Impact of water content on temporal evolution of chloride}
    \label{water}
    \vspace{-5mm}
\end{figure*}

Figure~\ref{water} illustrates the impact of concrete mixture water content on the temporal evolution of chloride using various ML algorithms---10 mm below the exposure surface. Given the simplicity of the LR algorithm, a linear response is captured by the LR method. The KNN algorithm also provides a simple representation of the overall trend, while other methods reveal a clearer pattern. In general, an increase in water content is found to increase chloride concentration over time for the LR, KNN, and MLP algorithms, while an inverse relationship is observed when employing the KRR, SVR, and GPR algorithms. Although an inconsistent relationship was found across different algorithms, the direct correlation seems more plausible when considering the physics of the problem.

\begin{figure*}[]
    \centering        
    \includegraphics[width=0.95\textwidth]{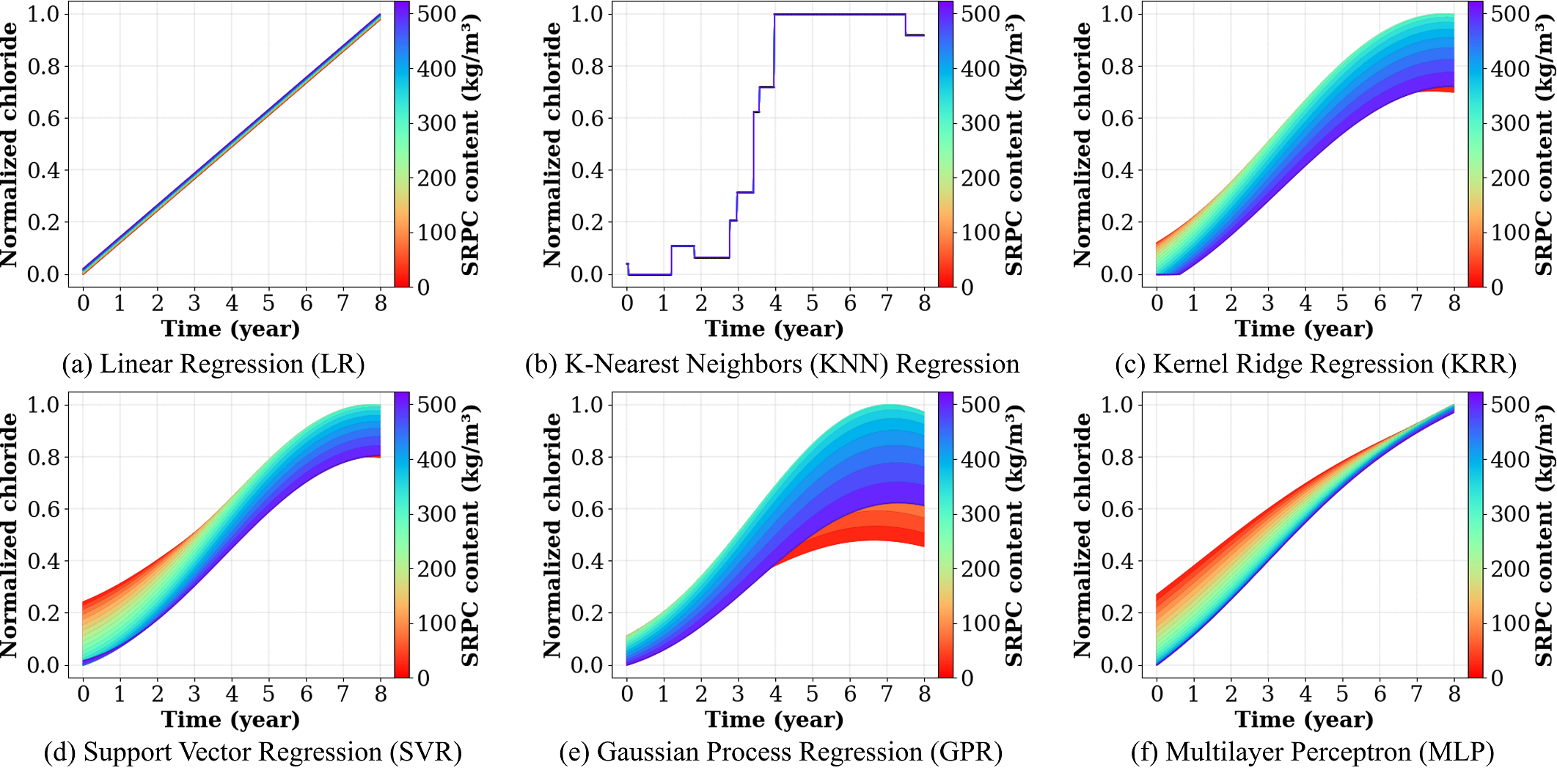}
    \vspace{-5mm}
    \caption{Impact of sulfate-resisting Portland cement (SPRC) content on temporal evolution of chloride}
    \label{SRPC}
    \vspace{-3mm}
\end{figure*}

The sensitivity of the results to the sulfate-resisting Portland cement (SPRC) content is depicted in Figure~\ref{SRPC}. Considering the predictive capability of each ML learner, a relatively clear trend is manifested by the MLP, wherein an increase SPRC content leads to a reduction in chloride content over time. Similar trends, to some extent, are also observed for the KRR, SVR, and GPR algorithms, indicating an inverse relationship between the temporal evolution of chloride and SPRC content.

\begin{figure*}[]
    \centering       
    \includegraphics[width=0.95\textwidth]{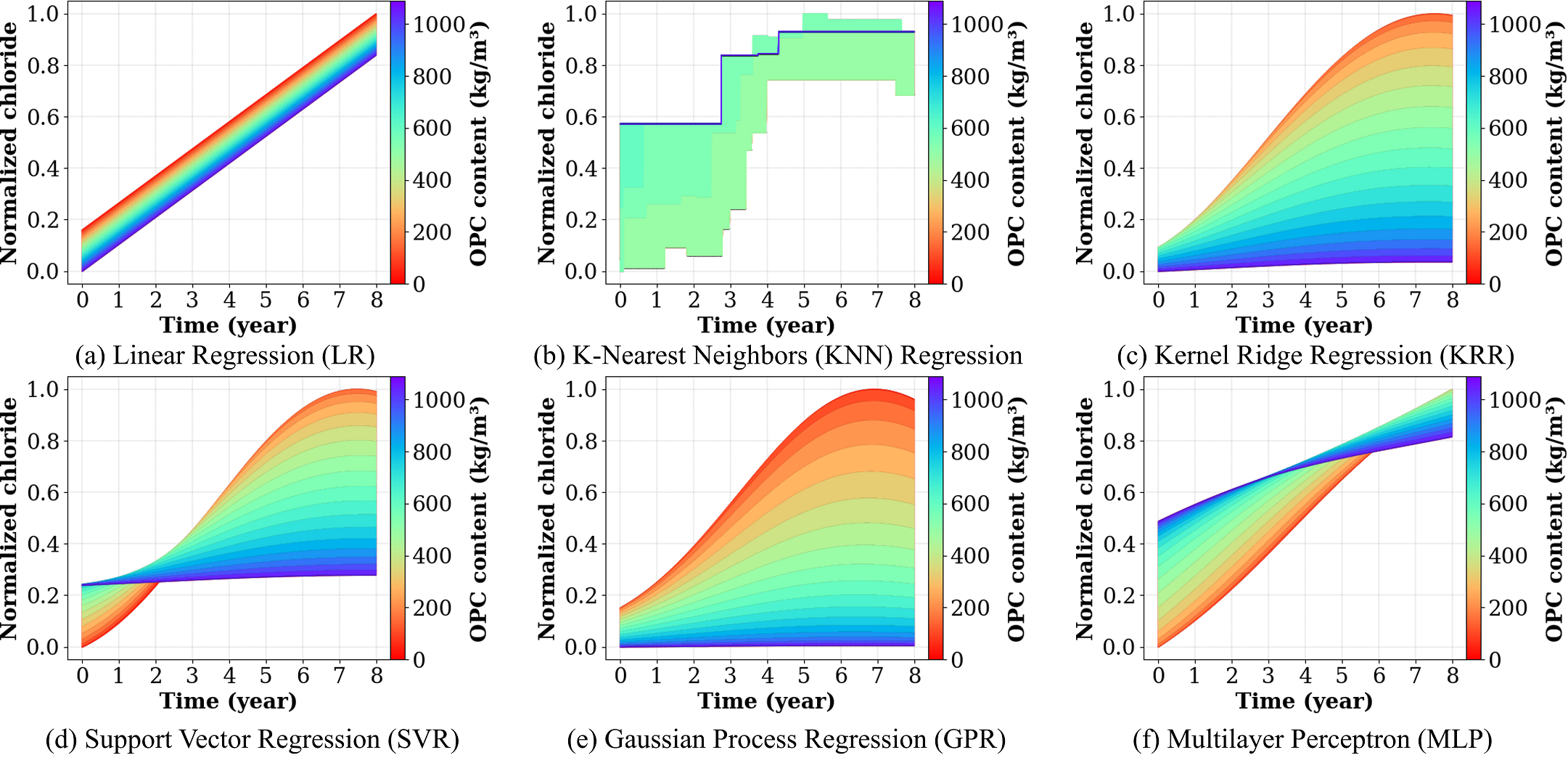}
    \vspace{-5mm}
    \caption{Impact of ordinary Portland cement (OPC) content on temporal evolution of chloride}
    \label{OPC}
    \vspace{-5mm}
\end{figure*}

The correlation between the content of ordinary Portland cement (OPC) and the temporal evolution of chloride is depicted in Figure~\ref{OPC}. Similar to preceding observations, the LR and KNN models provide simple estimates of the overall trend, while the KRR and GPR models manifest a clearer trend. In this respect, an increase in OPC content leads to a reduction in chloride concentration.

\begin{figure*}[]
    \centering        
    \includegraphics[width=0.95\textwidth]{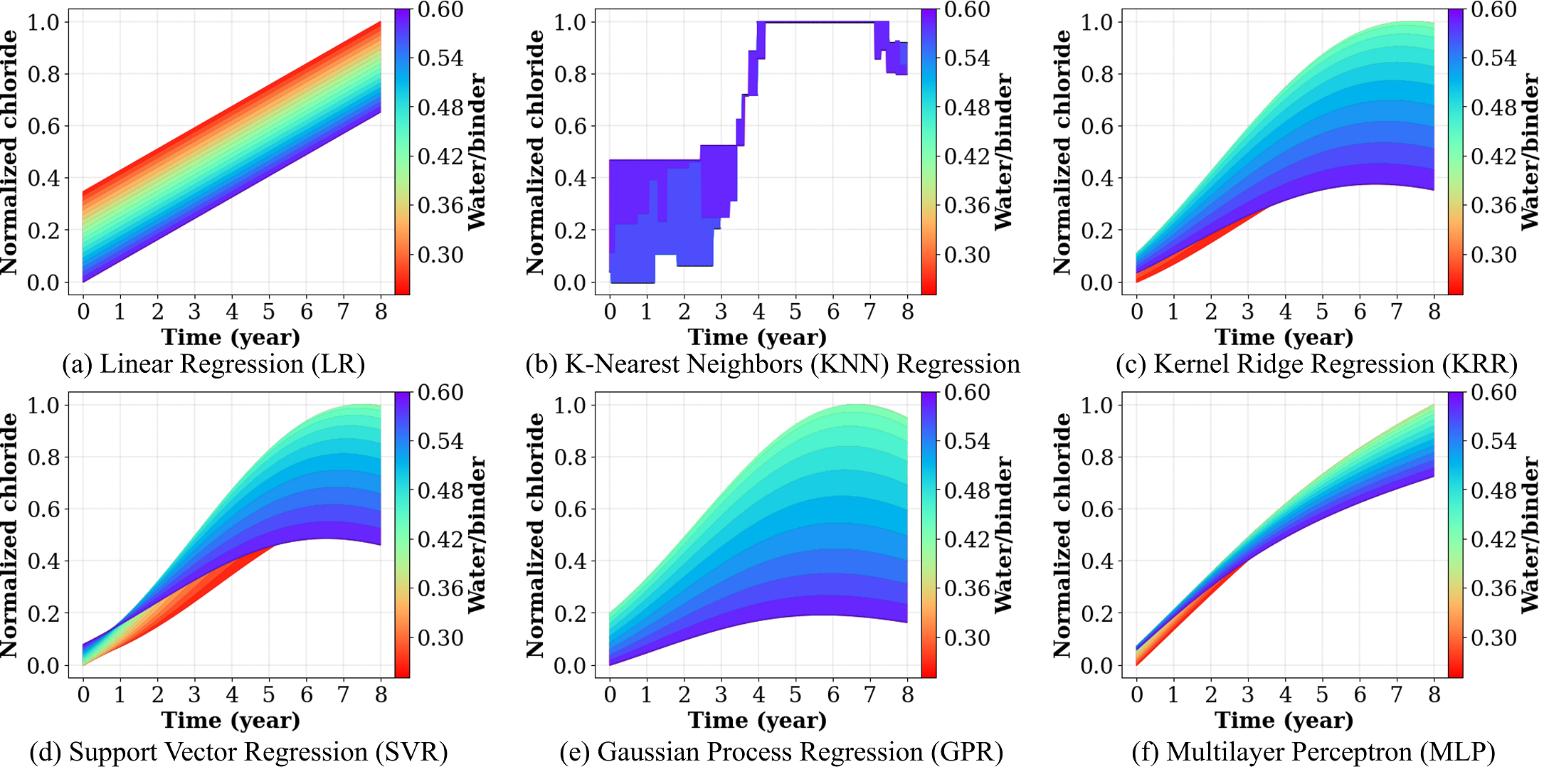}
    \vspace{-5mm}
    \caption{Impact of water-to-binder ratio on temporal evolution of chloride}
    \label{water-binder}
    \vspace{-3mm}
\end{figure*}

Figure~\ref{water-binder} shows the impact of the water-to-binder ratio on the temporal evolution of chloride. The simplified ML learners provide a rough estimate of this parameter. The advanced models provide a clearer picture of the overall trend; however, the trend seems to alternate over time. In this respect, in general, there exists a direct correlation between chloride content and water-to-binder ratio in the first few years, after which it changes to an inverse correlation.

\begin{figure*}[]
    \centering        
    \includegraphics[width=0.95\textwidth]{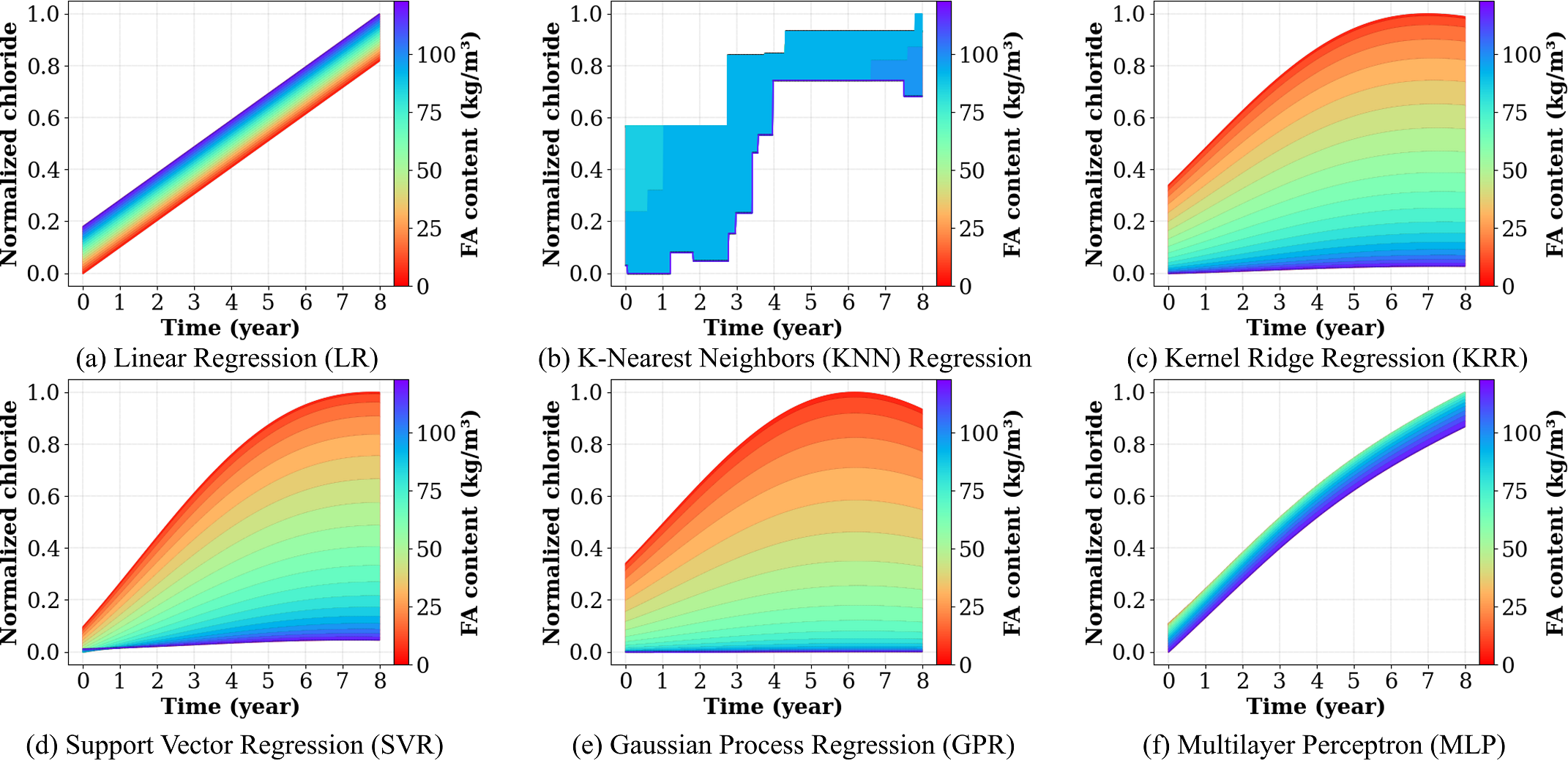}
    \vspace{-5mm}
    \caption{Impact of fly ash (FA) content on temporal evolution of chloride}
    \label{FlyAsh}
    \vspace{-5mm}
\end{figure*}

The impact of the fly ash (FA) content on the temporal evolution of chloride is depicted in Figure~\ref{FlyAsh}.  The sensitivity of the model response to this parameter is generally similar across all advanced ML algorithms. The relationship is particularly clear in the KRR, SVR, and GPR models, which demonstrate an inverse correlation, wherein an increase in FA content leads to a decrease in chloride concentration. 

\begin{figure*}[]
    \centering        
    \includegraphics[width=0.95\textwidth]{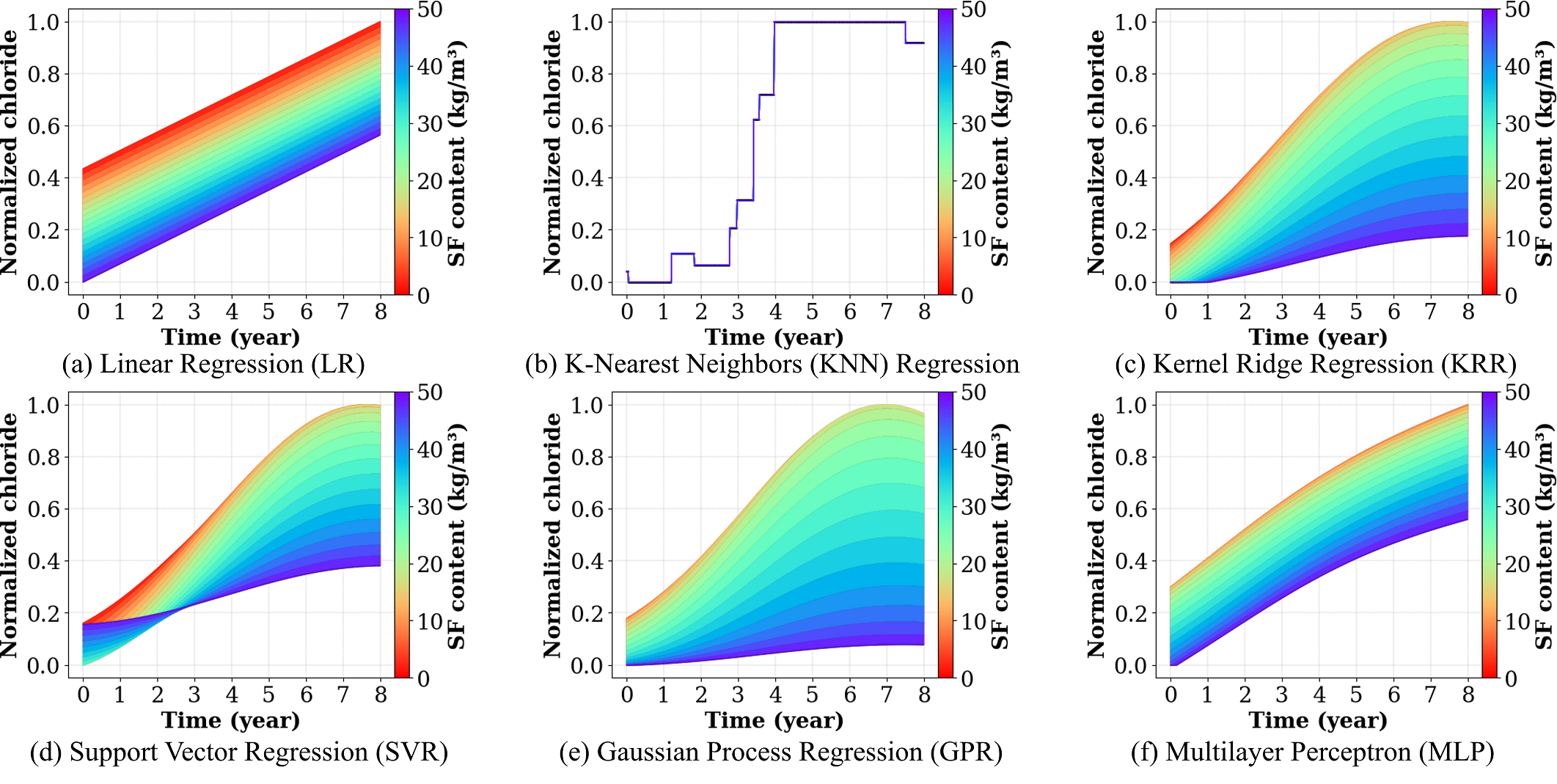}
    \vspace{-5mm}
    \caption{Impact of silica fume (SF) content on temporal evolution of chloride}
    \label{SilicaFume}
    \vspace{-3mm}
\end{figure*}

Figure~\ref{SilicaFume} demonstrates the correlation between silica fume (SF) content and chloride concentration over time. Among ML algorithms, the KRR, GPR, and MLP models capture a clear trend of the overall response. As the silica fume content increases, the concentration of chloride in the concrete decreases. As deduced from the figure, the optimum performance corresponds to the case with the highest SF content, which represents the most effective configuration for service life enhancement.

\begin{figure*}[]
    \centering        
    \includegraphics[width=0.95\textwidth]{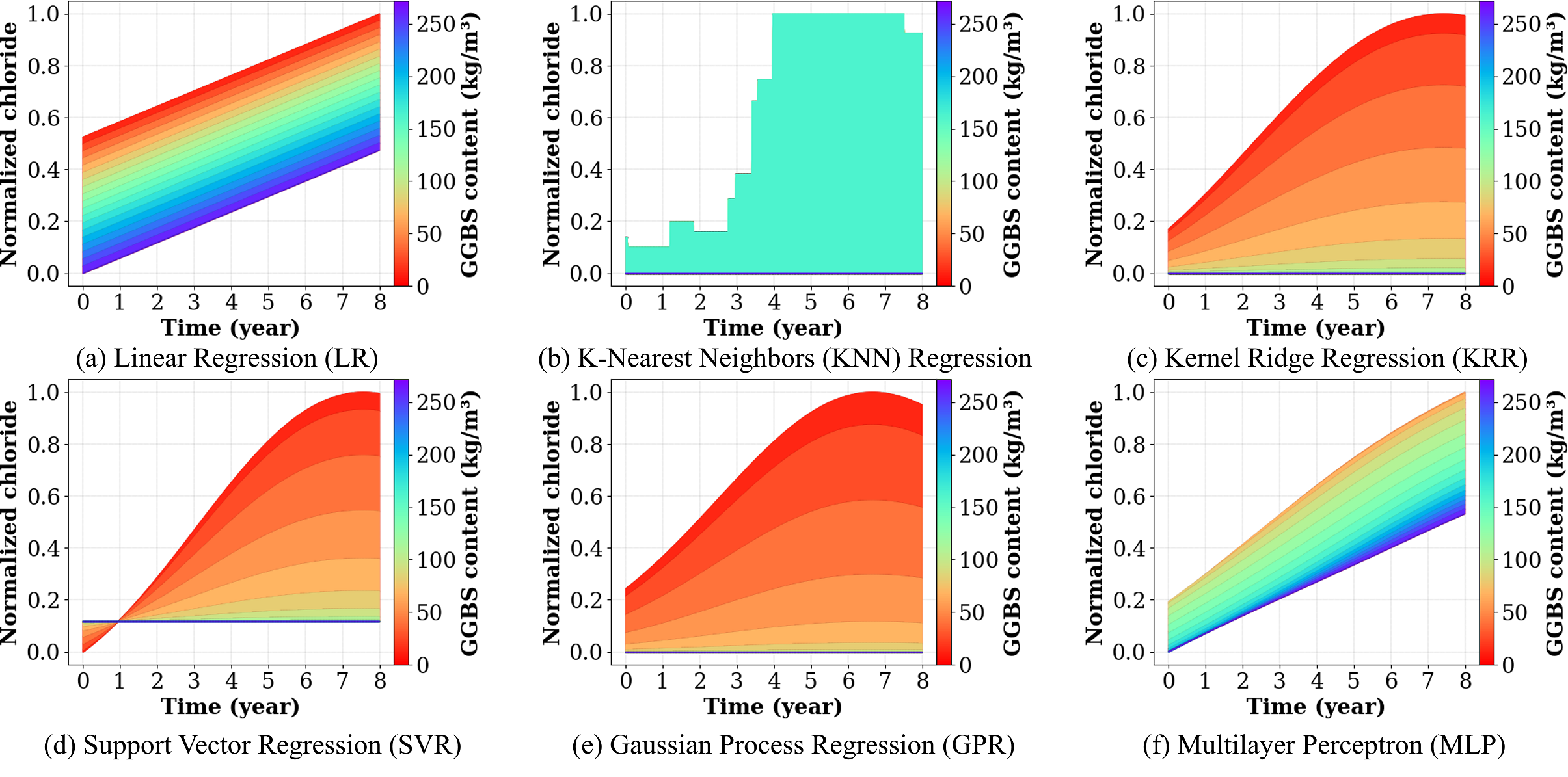}
    \vspace{-5mm}
    \caption{Impact of ground granulated blast-furnace slag (GGBS) content on temporal evolution of chloride}
    \label{GGBS}
    \vspace{-5mm}
\end{figure*}

The sensitivity of the results to the ground granulated blast-furnace slag (GGBS) content is illustrated in Figure~\ref{GGBS}. Consistent with previous observations, KRR and GPR provide a clear representation of the model response. In general, there exists an inverse correlation between high GGBS content and chloride concentration over time, where a significant increase in the former leads to a reduction in the latter; however, as deduced from the figure, chloride concentration is highly sensitive to GGBS content, with relatively small amounts producing a significant effect on chloride concentration over time. Similar trends are observed for SVR and, to some extent, the MLP.

\begin{figure*}[]
    \centering        
    \includegraphics[width=0.95\textwidth]{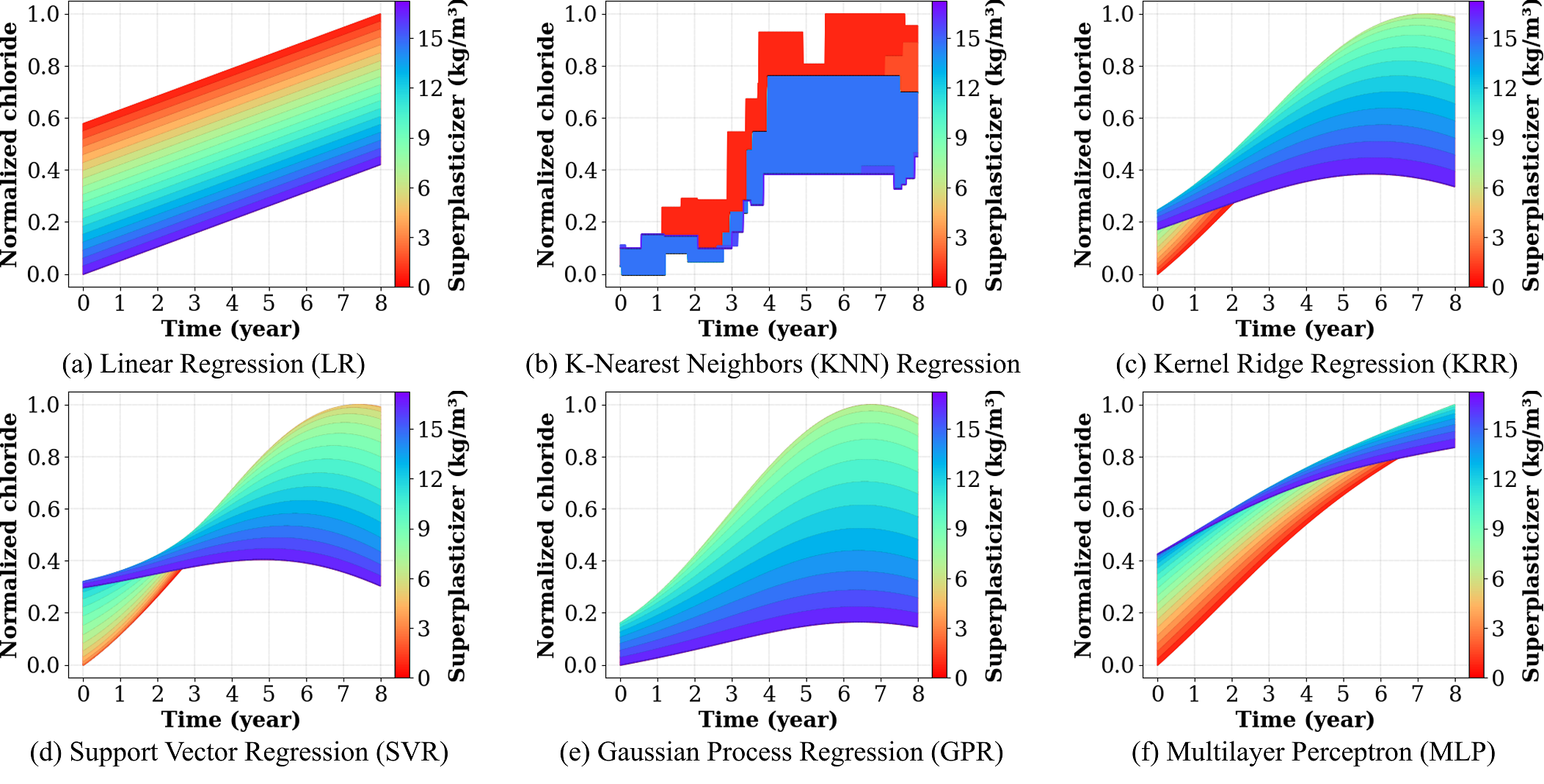}
    \vspace{-5mm}
    \caption{Impact of superplasticizer content on temporal evolution of chloride}
    \label{SuperPlasticizer}
    \vspace{-3mm}
\end{figure*}

Figure~\ref{SuperPlasticizer} shows the impact of superplasticizer on chloride concentration at a fixed point in time. Among the adopted ML algorithms, the GPR model provides a consistent trend, whereas the responses of other learners (SVR, KRR, and MLP) fluctuate over time. In this respect, an inverse relationship is predicted by GPR, while the MLP demonstrates a relatively direct correlation. The response of the KRR is also, to some extent, consistent with that of the GPR over time.

\begin{figure*}[]
    \centering        
    \includegraphics[width=0.95\textwidth]{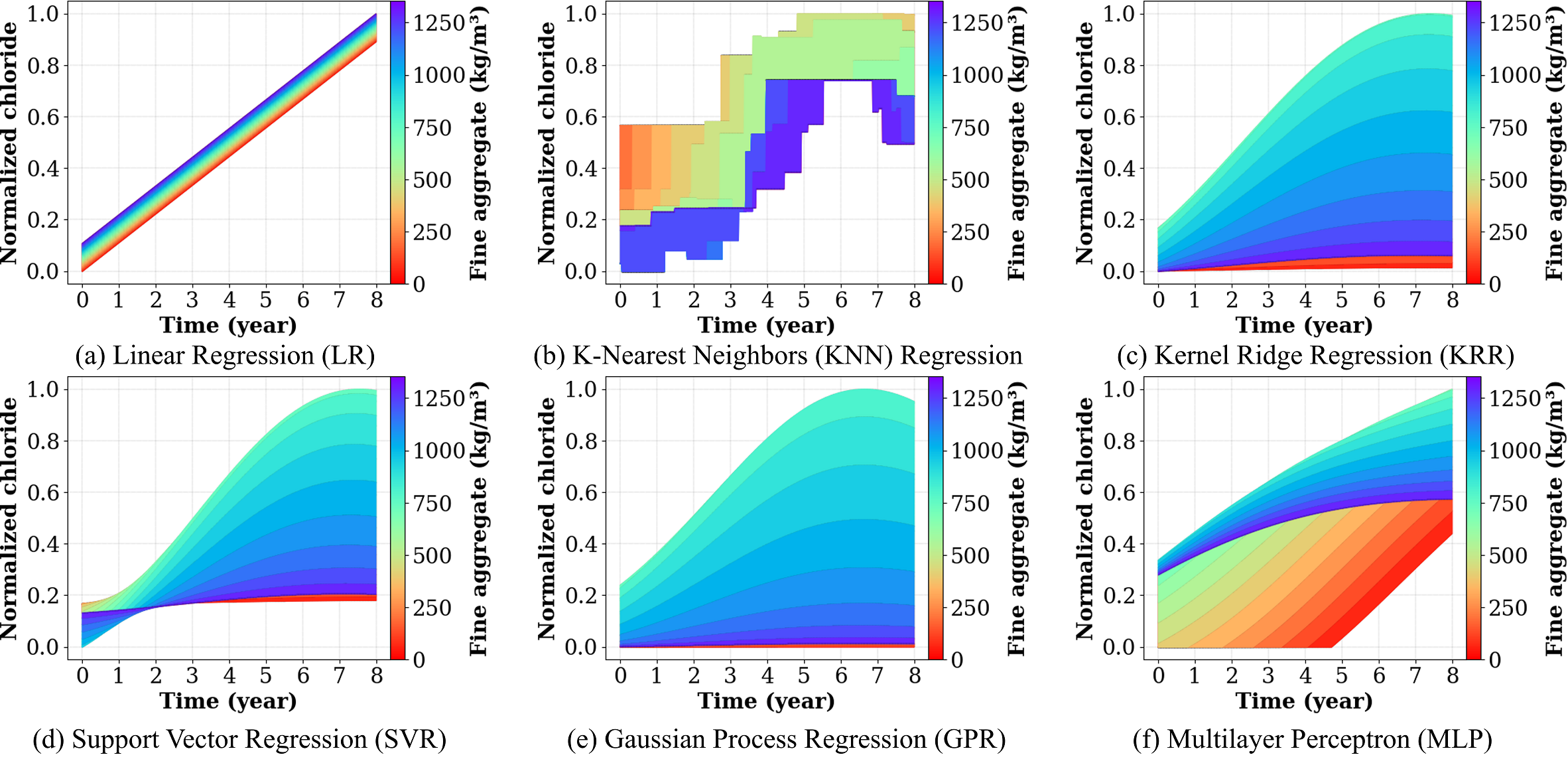}
    \vspace{-5mm}
    \caption{Impact of fine aggregate content on temporal evolution of chloride}
    \label{FineAggregate}
    \vspace{-5mm}
\end{figure*}

The correlation between fine aggregates and chloride content is further investigated and shown in Figure~\ref{FineAggregate}. Among the adopted ML algorithms, the responses obtained from KRR, SVR, and GPR are relatively similar. In general, an increase in fine aggregate content is found to reduce chloride concentration over time. However, the response is sensitive to the specific aggregate content. In particular, the response of MLP is clearer for lower aggregate contents. Considering the overall responses, there appears to be an optimal value of fine aggregate content---within the adopted range---that delivers the most promising response.

\begin{figure*}[]
    \centering        
    \includegraphics[width=0.95\textwidth]{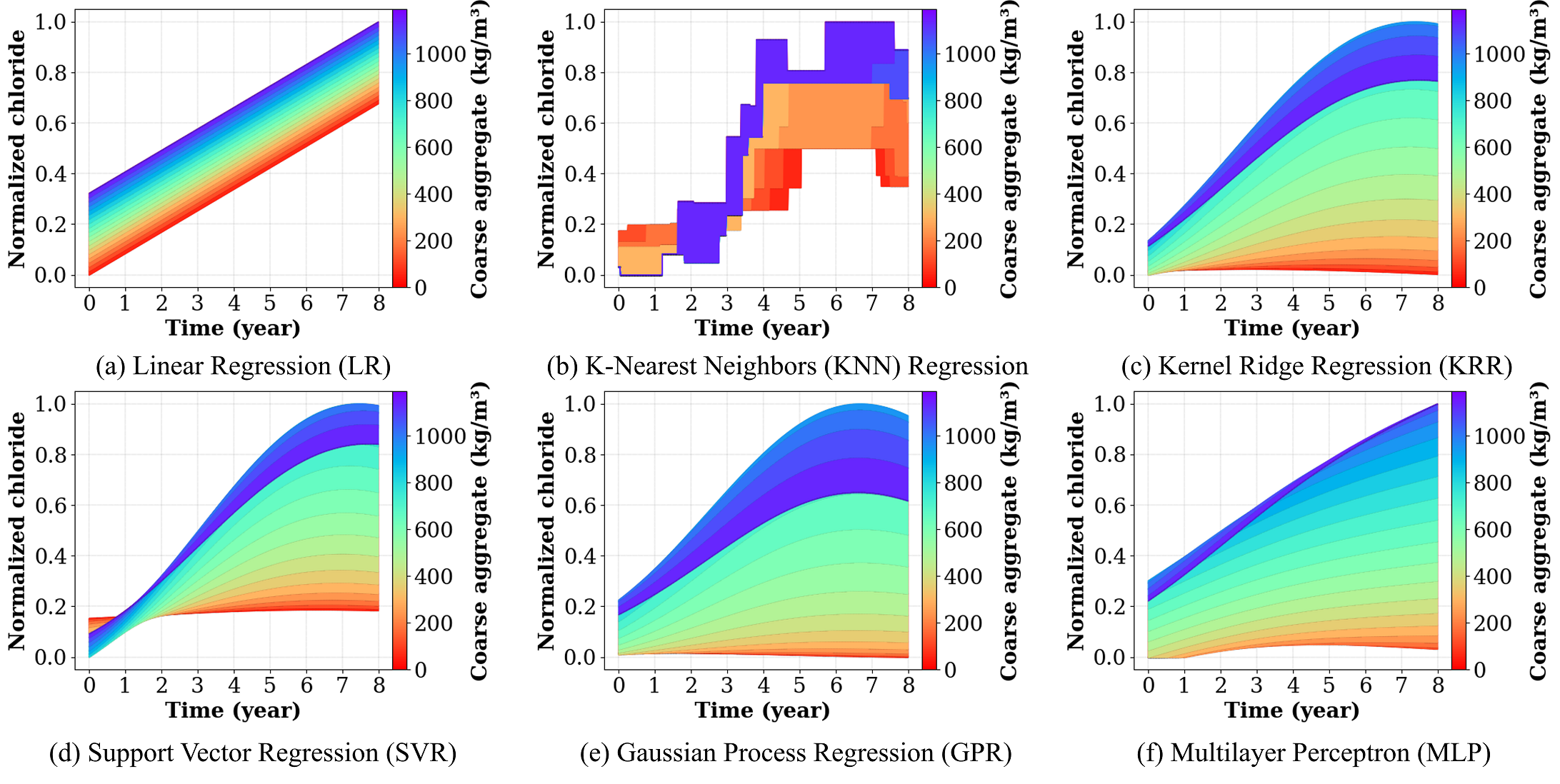}
    \vspace{-5mm}
    \caption{Impact of coarse aggregate content on temporal evolution of chloride}
    \label{CoarseAgg}
    \vspace{-5mm}
\end{figure*}

Subsequently, the correlation between coarse aggregate content and chloride concentration is depicted in Figure~\ref{CoarseAgg}. Among all ML algorithms, the KRR, GPR, and MLP models clearly reveal the underlying relationship, wherein an increase in coarse aggregate content leads to a corresponding increase in chloride concentration over time. However, there exists a specific quantity---just below the maximum adopted in this study---that produces the most adverse mixture, resulting in the highest chloride concentration. On the other hand, the optimal value for durability is associated with the minimum coarse aggregate content.

\section{Discussion}\label{Discussion}
%\vspace{0.5em} 
The evolution of chloride in concrete was successfully described using a variety of ML algorithms. The MLP delivered the most accurate results, achieving training and test scores of $R^2=0.94$ and $R^2=0.90$, respectively. The least accurate approaches were associated with the GRU and LR algorithms.

A significant observation pertains to the accuracy of the gated recurrent unit (GRU) network. As discussed in the foregoing sections, the fundamental premise of this approach lies in its ability to reproduce sequential data. Although the training set yielded a satisfactory performance with a score of $R^2=0.89$, the test set accuracy was poor ($R^2 = 0.55$). This discrepancy indicates that the sequential dependency present in the data imposes a significant constraint on the model, thereby compromising its predictive accuracy. The challenges in training and testing can be associated with the diversity of concrete mixture proportions adopted in the dataset.

An important observation regarding the hidden correlations between concrete mixture constituents and chloride ingress pertains to the ability of each ML learner to capture the latent relationships between input and output features. In this regard, LR provided a rough linear estimate of the underlying pattern. Similarly, KNN produced a rather discrete and somewhat unclear representation. SVR and MLP captured a clear trend for the majority of mixture compositions, although in some cases they exhibited variations in the temporal pattern. The KRR also delivered an acceptable response, with a similar limitation, while GPR consistently produced a clear trend across the entire input features.

Pursuant to the assessment of all ML algorithms, the hidden correlation between input and output features was unraveled. An inverse relationship was observed between the temporal evolution of chloride content and several concrete mixture compositions, including sulfate-resisting Portland cement, ordinary Portland cement, fly ash, silica fume, and superplasticizer. A similar trend was relatively found for fine aggregate content. In contrast, a clear direct correlation was identified between chloride content and coarse aggregate content, whereby an increase in the latter resulted in an increase in the former. The correlations of chloride content with water content and the water-to-binder ratio were not very clear and alternated over time; however, the MLP provided a general direct correlation with respect to these features in the early stages of exposure. The evolution of chloride was highly sensitive to ground granulated blast-furnace slag (GGBS), with small variations in its quantity leading to substantial changes in the chloride content.

A noteworthy consideration pertains to the physical interpretation of the observed phenomena. While the adopted ML algorithms provide insight into the relationships between input and output features, the underlying physical mechanisms governing these observations require further investigation and constitute a topic for future study. In this regard, some of the observed trends may be partially explained by changes in the pore morphology of concrete over time. For example, an increase in coarse aggregate content may lead to higher overall porosity, which in turn facilitates chloride diffusion. In contrast, increasing the fly ash content can reduce concrete porosity and, consequently, diminish the temporal diffusion of chloride. In addition, secondary chemical reactions associated with mixture compositions may influence the temporal evolution of chloride ingress. A detailed investigation of these mechanisms, which is beyond the scope of the present study, is reserved for future research.

The description of the hidden correlations was successfully achieved using a variety of ML algorithms. However, the optimization of concrete mixture proportions represents an interesting direction for future research. Such investigations may presently be challenging due to the limited number of experimental measurements available in the literature, as well as the complex coupling among input features and space-time domain.

\section{Conclusions}\label{Conclusion}
This study employs various machine learning (ML) algorithms to assess the impact of concrete mixture compositions on chloride ingress in concrete structures. The adopted standalone ML algorithms consist of three simple and four complex methods. The simple algorithms include linear regression (LR), k-nearest neighbors (KNN) regression, and kernel ridge regression (KRR). The complex algorithms comprise support vector regression (SVR), Gaussian process regression (GPR), and two families of artificial neural networks, including multilayer perceptron (MLP) and gated recurrent unit (GRU). The KRR, GPR, and MLP algorithms successfully predicted the chloride content in concrete, achieving a minimum test-set score of $R^2=0.90$. The GRU was unable to accurately capture the sequential response in the test set, while achieving a good score on the training set. The latent relationships between the input features and the temporal evolution of chloride were appraised. The LR and KNN algorithms provided a rough estimate, while SVR, MLP, and KRR provided an acceptable and explainable response. On the other hand, GPR demonstrated the most promising performance by capturing clear and explainable trends. A clear inverse correlation was observed between chloride content and several mixture compositions, including the contents of sulfate-resisting Portland cement, ordinary Portland cement, fly ash, silica fume, and superplasticizer, as well as, to a lesser extent, fine aggregate. In contrast, a clear direct correlation was identified between coarse aggregate content and chloride ingress. The results were found to be sensitive to the content of ground-granulated blast-furnace slag. These findings emphasize the very premise of surrogate approaches for describing chloride ingress in concrete, while concurrently identifying the latent correlations toward the ultimate goal of extending the service life of civil infrastructure.

% To print the credit authorship contribution details
\printcredits

\section*{Declaration of competing interest}
The authors declare that they have no known competing financial
interests or personal relationships that could have appeared to influence the work reported in this paper.

%\section*{Acknowledgment}
%The authors did not receive funding for %this work. Any opinions expressed in this %paper are those of the authors alone.

%% Loading bibliography style file
%\bibliographystyle{model1-num-names}
%\bibliographystyle{cas-model2-names}
\bibliographystyle{elsarticle-num}
%\bibliographystyle{elsarticle-harv}
%\bibliographystyle{unsrt}  % Bibliography style that orders references by citation order

%\bibliographystyle{plain}
% Loading bibliography database
\bibliography{cas-refs}

% Biography
%\bio{}
% Here goes the biography details.
%\endbio

%\bio{pic1}
% Here goes the biography details.
%\endbio

\end{document}